\documentclass[11pt]{article}

\usepackage[preprint]{acl}

\usepackage{times}
\usepackage{latexsym}

\usepackage[T1]{fontenc}
\usepackage[utf8]{inputenc}
\usepackage{microtype}
\usepackage{inconsolata}

\usepackage{graphicx}
\usepackage{amsmath}
\usepackage{booktabs}
\usepackage{multirow}
\usepackage[table]{xcolor}
\usepackage{array}
\usepackage{makecell}

\usepackage{booktabs}
\usepackage{multirow}
\usepackage{makecell}
\usepackage[table]{xcolor}
\usepackage{graphicx}
\usepackage{array}
\usepackage{tcolorbox}
\tcbuselibrary{breakable}

\title{Med-Banana: Learning Quality-Controlled Medical Image Editing from Success-and-Failure Trajectories}

\author{
 \textbf{Zhihui Chen\textsuperscript{1}},
 \textbf{Qingyuan Lei\textsuperscript{2}},
 \textbf{Kai He\textsuperscript{1}},
 \textbf{Yanrui Du\textsuperscript{3}},
 \textbf{Mengling Feng\textsuperscript{1}\thanks{Corresponding author}}
\\
\\
\textsuperscript{1}National University of Singapore \\
\textsuperscript{2}The Chinese University of Hong Kong 
\textsuperscript{3}Harbin Institute of Technology
\\
 \texttt{zhihui.chen@u.nus.edu, \{kai\_he, ephfm\}@nus.edu.sg}\\
 \texttt{qingyuan.lei@link.cuhk.edu.hk}\\
}

\begin{document}
\maketitle

\begin{abstract}
Text-guided medical image editing must satisfy the requested pathology while preserving anatomy, modality-specific appearance, and clinical plausibility. However, existing datasets largely supervise editors with final accepted edits and discard the failed attempts produced during generation. We argue that these failures provide essential supervision for quality control: they specify what should be rejected, why an edit is medically or visually invalid, and how the instruction should be revised. We present \textbf{Med-Banana}, a trajectory-supervised framework for quality-controlled medical image editing. We introduce \textbf{Med-Banana-80K}, a large-scale resource of success-and-failure editing trajectories with candidate images, verification outcomes, rejection reasons, and prompt refinements. Building on it, Med-Banana jointly trains an editor, verifier, and refiner, enabling edit--verify--refine inference from accepted and rejected attempts. Experiments across MLLM judges, blind expert assessment, source-preservation and real--synthetic separability probes demonstrate consistent improvements over open medical image editors. Code and data are publicly available\footnote{\url{https://anonymous.4open.science/r/Med-Banana-anonymous-E1E7/}}.
\end{abstract}

\section{Introduction}

Text-guided medical image editing aims to modify clinical images according to natural-language instructions while preserving the medically relevant content of the source image. This capability supports controllable medical visual data construction, medical imaging model development, and counterfactual medical image analysis~\cite{selfimprove_gen_medical,prism}. Unlike general image editing, however, medical image editing is constrained by domain-specific validity. A qualified edit should introduce the requested pathology without disrupting patient-specific anatomy, modality-specific appearance, disease plausibility, or image-formation characteristics~\cite{medforge}. Medical image editing is therefore not only a conditional generation task, but also a quality-controlled editing problem.

Existing medical image editing datasets and editors are largely built on final accepted edits~\cite{med-edit-gan, MedEBench, miedb-bench, mededit}. This supervision format provides direct examples of successful editing outcomes, but removes the generation process that precedes them. In practice, many candidates fail before an acceptable edit is obtained. They may miss the target finding, alter non-target anatomy, introduce clinically implausible patterns, or produce modality-inconsistent artifacts~\cite{radedit}. These rejected attempts are typically excluded from training data, together with the diagnostic feedback and prompt revisions generated during quality control.

This omission limits what medical editing models can learn from existing datasets. Final accepted edits mainly supervise endpoint imitation: they show what a qualified output may look like, but provide little information about the boundary between acceptable and invalid edits. Rejected candidates contain complementary supervision. They expose failure modes, associate visual evidence with medical or imaging constraints, and record how an instruction should be revised after an unsuccessful attempt. Such information is particularly relevant in medical imaging, where an edit may appear visually plausible while remaining clinically inconsistent~\cite{align_synth_med}.

We introduce \textbf{Med-Banana}, a trajectory-supervised framework for quality-controlled medical image editing. Its core resource, \textbf{Med-Banana-80K}, preserves the edit--verify--refine process rather than retaining only final accepted outputs. Each trajectory records the source image, target disease, candidate edit, verification outcome, rejection reason when the candidate fails, and prompt refinement for the next attempt. This formulation changes the supervision unit from isolated final edits to success-and-failure editing trajectories.

Med-Banana uses these trajectories to train a coupled editing system. Accepted trajectories provide positive supervision for adapting the image editor, while rejected trajectories supervise candidate verification and prompt refinement. At inference time, the system performs edit--verify--refine from the original source image: it generates a candidate edit, verifies whether the result satisfies medical and visual constraints, and revises the prompt when verification fails. Editing from the original image at each retry avoids artifact accumulation while allowing verifier feedback to guide subsequent attempts.

We evaluate Med-Banana with automatic medical judges, blind expert assessment, and objective probes for source preservation and real--synthetic separability. Across these protocols, Med-Banana consistently improves open medical image editors, demonstrating the value of modeling rejected candidates and refinement traces as first-class supervision for medical image editing.

The main contributions are as follows:
\begin{itemize}
    \item We introduce \textbf{Med-Banana-80K}, a large-scale medical image editing trajectory dataset that preserves accepted edits, rejected attempts, verification outcomes, rejection reasons, and prompt refinement traces.
    \item We formulate \textbf{success-and-failure trajectories} as a supervision unit for medical image editing, enabling positive learning from accepted edits and negative/corrective learning from failed attempts.
    \item We instantiate \textbf{Med-Banana} as a quality-controlled edit--verify--refine system, improving open editors by up to $+1.27$ under MLLM Judge and $+0.84$ under blind expert assessment, and achieving the best overall rank ($1.50$) on source-preservation and real--synthetic separability probes.
\end{itemize}

\begin{table}[t]
\centering
\resizebox{\linewidth}{!}{%
\setlength{\tabcolsep}{4pt}
\renewcommand{\arraystretch}{1.15}
\begin{tabular}{l c l}
\toprule
\textbf{Resource} & \textbf{Size} & \textbf{Exposed Supervision} \\
\midrule
MedEdit~\cite{mededit} & $443$ & Final success pairs \\
MedE-Bench~\cite{MedEBench} & $\sim$1K & Final success pairs \\
RadEdit~\cite{radedit} & $\sim$6K & Final success pairs \\
MedGEN-Bench~\cite{medgen-bench} & $\sim$6K & Final success pairs \\
MieDB-100k~\cite{miedb-bench} & $\sim$100K & Success editing triplets \\
Med-Banana (Ours) & $\sim$80K & Success \& Failure Traj. \\
\bottomrule
\end{tabular}
}
\caption{\textbf{Medical image editing resources.}
Existing resources expose accepted edits as supervision, whereas Med-Banana retains accepted and rejected attempts with verification and refinement traces.}
\label{tab:dataset_comparison}
\end{table}

\section{Related Work}

\paragraph{Medical Image Editing Methods, Benchmarks, and Datasets.}
Medical image editing has evolved from task-specific pathology manipulation to broader text-guided editing resources. MedEdit and PRISM study diffusion-based editing that modifies disease-related content while preserving non-target structures~\cite{mededit,prism}. RadEdit creates controlled chest X-ray variations for stress-testing biomedical vision models, emphasizing the need to avoid spurious global artifacts~\cite{radedit}. Recent benchmarks further expand evaluation and coverage: MedE-Bench provides clinically curated image-prompt pairs, ROI annotations, and diagnostic evaluation dimensions~\cite{MedEBench}; MedGEN-Bench situates image editing within multimodal medical generation~\cite{medgen-bench}; and MieDB-100k scales editing data across perception, modification, and transformation tasks~\cite{miedb-bench}. As summarized in Table~\ref{tab:dataset_comparison}, these resources improve the scale and evaluation of medical editing, but their learnable supervision remains centered on accepted final edits. Med-Banana targets the missing process signal: rejected candidates, verification outcomes, failure reasons, and refinement traces.

\noindent\textbf{Quality-Controlled and Feedback-Driven Image Editing.} General-domain editing increasingly treats image editing as an iterative or agentic process. RefineEdit studies feedback-driven refinement~\cite{RefineEdit}. CAMEO embeds evaluation and revision in a multi-agent editing loop~\cite{quality-image-edit}, while ImageEdit-R1 formulates multi-agent editing as sequential decision making with reinforcement learning~\cite{image-edit-r1}. These works motivate feedback-based controllability, but they are not designed around medical validity or failure supervision. Med-Banana focuses on medical editing, where verification accounts for instruction compliance, pathological plausibility, anatomical consistency, and imaging fidelity. It further stores rejected candidates, verifier rationales, and prompt refinements as trajectory-level supervision for verification and feedback-guided refinement.

\begin{figure*}[t]
    \centering
    \includegraphics[width=\linewidth]{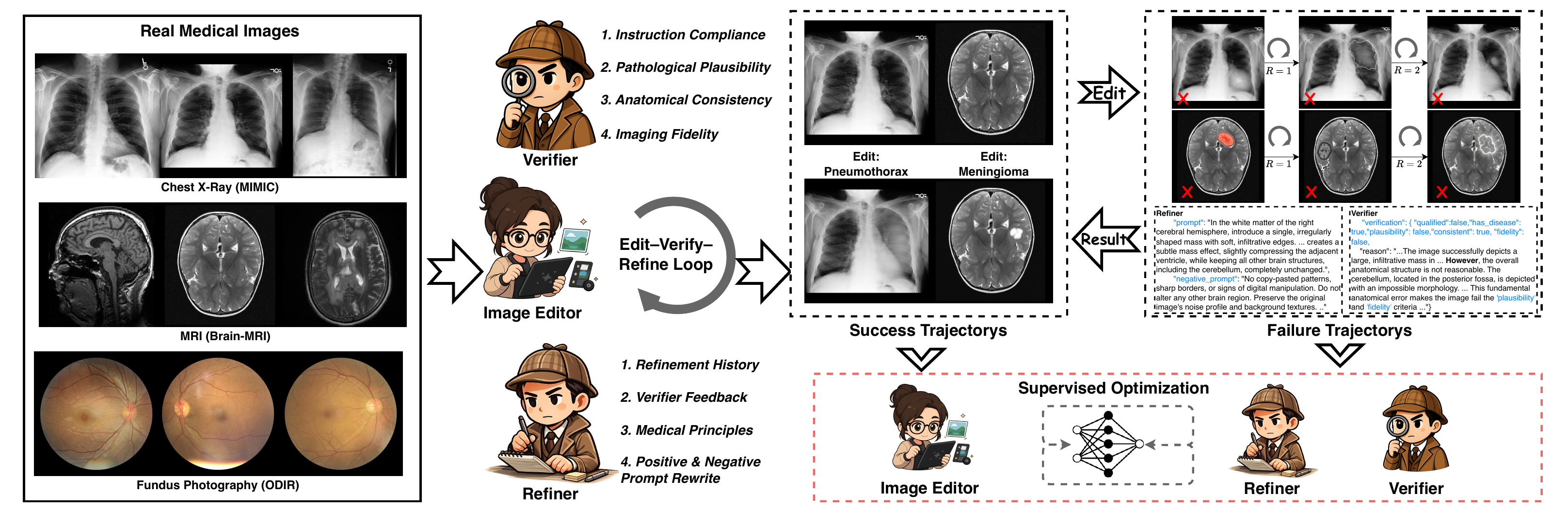}
\caption{\textbf{Overview of Med-Banana.}
Med-Banana builds success-and-failure editing trajectories through an edit--verify--refine loop. Accepted candidates supervise editor adaptation; rejected candidates are stored with verifier rationales and prompt refinements. These trajectories turn accepted edits into positive supervision, and rejected attempts into diagnostic and corrective supervision for verification and feedback-guided refinement.}
    \label{fig:system_overview}
\end{figure*}

\section{Med-Banana-80K: Success and Failure Trajectories of Medical Image Editing}
\label{sec:medbanana80K}

We introduce Med-Banana-80K, a large-scale medical image editing dataset built around success-and-failure trajectories. Unlike final-pair datasets that retain only accepted outputs, Med-Banana-80K preserves the edit--verify--refine process. Each trajectory records the edited image, verification outcome, rejection reason and instruction refinement.

\paragraph{Source Images and Disease Coverage.}
We collect 25,636 real clinical images from three representative 2D medical imaging modalities: chest X-ray, brain MRI, and fundus photography. The images are drawn from MIMIC~\cite{mimic}, Brain-MRI~\cite{mri}, and ODIR~\cite{odir}, covering 23 pathological categories across thoracic findings, brain tumors, and retinal diseases. These modalities impose distinct editing constraints: chest X-ray edits require global radiographic consistency~\cite{radedit}, brain MRI edits require preserving anatomical structures around tumour regions~\cite{latent_brain}, and fundus edits must maintain lesion morphology, vessel continuity, and local retinal texture~\cite{generation_fundus}.

\paragraph{Trajectory Construction.}
We construct trajectories through the edit--verify--refine loop in Figure~\ref{fig:system_overview}. Given a source image and target disease, the editor generates a candidate image from the current editing prompt and negative prompt. The verifier evaluates the candidate using four criteria: \textit{instruction compliance}, \textit{pathological plausibility}, \textit{anatomical consistency}, and \textit{imaging fidelity}. Accepted candidates are stored as success trajectories. Rejected candidates are stored with verifier reasons, which are then passed to the refiner to produce a revised prompt pair. Each new round edits from the original source image rather than the rejected candidate, avoiding artifact accumulation while preserving the full sequence of attempts.

\paragraph{Trajectory Representation.}
Each trajectory is represented as a round-indexed record of the interaction among editor, verifier, and refiner. At round $r$, the record contains the source image, target disease, editing prompt, negative prompt, candidate image, verifier decision, and verifier rationale. If the candidate is rejected, the record additionally stores the refined prompt and negative prompt for round $r+1$. This format specifies what was attempted, why the attempt failed or succeeded, and how the next attempt was revised.

\paragraph{Success and Failure Supervision.}
Med-Banana-80K provides two complementary supervision signals. Success trajectories provide positive source--instruction--edited-image triples for adapting medical image editors. Failure trajectories provide rejected candidates with explicit medical rejection reasons and refinement actions. These failed attempts support recognizing unsatisfactory edits, explaining failure modes, and generating follow-up instructions. Thus, Med-Banana-80K serves not only as an editing dataset, but also as supervision for verification and feedback-guided refinement.

\paragraph{Dataset Scale.}
Med-Banana-80K contains 25,636 original images across chest X-ray, brain MRI, and fundus photography, spanning 23 pathological categories. It includes 50,635 success trajectories and 37,822 failure trajectories. The released corpus contains 114,093 image records in total, with 88,457 success-and-failure editing attempts forming the core trajectory supervision.

\section{Med-Banana System}
\label{sec:methodology}

In this section, we propose Med-Banana, a quality-controlled medical image editing system based on success-and-failure trajectories. It defines the edit-addition task, describes trajectory-based supervision, and presents the edit--verify--refine loop.

\subsection{Quality-Controlled Editing Formulation}
Given a source medical image $\boldsymbol{x}$, a target disease $d$, and an initial editing instruction $p_0$, the system generates a candidate edited image $\hat{\boldsymbol{x}}$. In the main experiments, the initial instruction follows the same concise template for all samples: \textit{add \{disease\} to this image}. This simple prompt intentionally provides only the target condition, to measure the benefit of trajectory-guided refinement.

Med-Banana decomposes the editing process into three modules. The \textit{editor} maps the source image and current prompt pair to a candidate edit. The \textit{verifier} compares the original and edited images and determines whether the candidate satisfies instruction compliance, pathological plausibility, anatomical consistency, and image-physics fidelity. The \textit{refiner} converts the verifier reason and previous failure history into a revised prompt and negative prompt for the next attempt.

\subsection{Learning from Success and Failure}
\noindent\textbf{Image editor tuning.}
Successful trajectories provide positive editing supervision for the image editor. Each accepted attempt gives a source image, an editing instruction, and an accepted edited image. We use these triples to adapt open image editors with LoRA tuning. This keeps the editor training close to conventional supervised editing, while grounding the supervision in trajectory records whose rejected attempts are also preserved.

\noindent\textbf{Verifier tuning.}
Success and failure trajectories provide supervision for candidate assessment. We convert each edited image into an inquiry--response pair for a vision-language verifier. The inquiry asks whether the candidate satisfies the requested pathology and remains structurally reasonable and visually realistic. The response contains a structured quality decision and a concise reason, covering instruction compliance, pathological plausibility, anatomical consistency, and imaging fidelity. Successful attempts teach the verifier what should pass, while failed attempts teach which visual or medical failures should be rejected.
\begin{table*}[ht]
\centering
\large
\resizebox{\linewidth}{!}{%
\setlength{\tabcolsep}{6pt}
\begin{tabular}{l|cc|cc|cc|cc|cc}
\toprule
\multirow{2}{*}{\textbf{Model}}
& \multicolumn{2}{c|}{\textbf{Instruction Compliance}}
& \multicolumn{2}{c|}{\textbf{Pathological Plausibility}}
& \multicolumn{2}{c|}{\textbf{Anatomical Consistency}}
& \multicolumn{2}{c|}{\textbf{Imaging Bio-Fidelity}}
& \multicolumn{2}{c}{\textbf{Overall}} \\
\cmidrule(lr){2-3}
\cmidrule(lr){4-5}
\cmidrule(lr){6-7}
\cmidrule(lr){8-9}
\cmidrule(lr){10-11}
& \textbf{\quad Mean \quad} & \textbf{ Std. \quad}
& \textbf{\quad Mean \quad} & \textbf{Std. \quad}
& \textbf{\quad Mean \quad} & \textbf{Std. \quad}
& \textbf{\quad Mean \quad} & \textbf{Std. \quad}
& \textbf{Average} & \textbf{Rank} \\
\midrule

\rowcolor{blue!15}
\multicolumn{11}{l}{\textbf{MiMo-V2.5 Judge}} \\
GPT-Image-2                    & \textbf{8.23} & 0.62 & \textbf{7.37} & 0.90 & 8.47 & 0.52 & 7.95 & 1.03 & \textbf{8.01} & \textbf{1} \\
Nano-Banana-2                  & \underline{8.13} & 0.76 & \underline{7.27} & 1.06 & 8.51 & 0.50 & 7.63 & 1.26 & 7.89 & 4 \\
Nano-Banana                    & 8.07 & 0.69 & 7.12 & 1.03 & 8.52 & 0.50 & 7.82 & 1.14 & 7.88 & 5 \\
Qwen-Image-Edit-2512           & 7.95 & 0.65 & 7.00 & 1.01 & 8.46 & 0.50 & 7.65 & 1.14 & 7.77 & 6 \\
FireRed-Image-Edit-1.1         & 7.91 & 0.70 & 7.06 & 1.02 & 8.36 & 0.48 & 7.44 & 1.04 & 7.69 & 7 \\
\rowcolor{gray!15}
Med-Banana\,(Qwen)             & 8.07 & 0.70 & 7.06 & 1.07 & \textbf{8.57} & 0.50 & \underline{8.02} & 1.02 & \underline{7.93} & \underline{2} \\
\rowcolor{gray!15}
Med-Banana\,(FireRed)          & 8.03 & 0.67 & 7.03 & 1.02 & \underline{8.56} & 0.52 & \textbf{8.11} & 0.93 & \underline{7.93} & \underline{2} \\
\midrule
\rowcolor{blue!15}
\multicolumn{11}{l}{\textbf{MedGemma-27B Judge}} \\
GPT-Image-2                    & \textbf{5.76} & 2.10 & \underline{3.84} & 1.73 & \textbf{6.47} & 2.32 & \textbf{3.84} & 1.50 & \textbf{4.98} & \textbf{1} \\
Nano-Banana-2                  & \underline{5.75} & 2.17 & \textbf{3.92} & 1.94 & 5.90 & 2.81 & \underline{3.71} & 1.82 & \underline{4.82} & \underline{2} \\
Nano-Banana                    & 4.57 & 1.72 & 2.81 & 1.09 & 5.52 & 2.87 & 3.09 & 1.84 & 4.00 & 4 \\
Qwen-Image-Edit-2512           & 3.28 & 1.80 & 1.97 & 1.11 & 4.71 & 3.20 & 2.19 & 1.78 & 3.04 & 6 \\
FireRed-Image-Edit-1.1         & 2.97 & 1.85 & 1.78 & 1.18 & 3.72 & 3.19 & 2.34 & 2.40 & 2.70 & 7 \\
\rowcolor{gray!15}
Med-Banana\,(Qwen)             & 4.27 & 1.72 & 2.72 & 1.13 & \underline{6.19} & 2.58 & 3.21 & 1.85 & 4.10 & 3 \\
\rowcolor{gray!15}
Med-Banana\,(FireRed)          & 4.47 & 1.52 & 2.73 & 1.01 & 5.86 & 2.56 & 2.84 & 1.28 & 3.97 & 5 \\

\midrule
\rowcolor{blue!15}
\multicolumn{11}{l}{\textbf{Expert as Judge}} \\
GPT-Image-2                    & \textbf{7.00} & 2.17 & \textbf{6.50} & 2.03 & \textbf{7.07} & 1.79 & \textbf{7.07} & 2.02 & \textbf{6.91} & \textbf{1} \\
Nano-Banana-2                  & \underline{6.72} & 2.74 & \underline{6.44} & 2.54 & \underline{6.61} & 2.00 & \underline{7.00} & 2.11 & \underline{6.69} & \underline{2} \\
Nano-Banana                    & 5.63 & 1.74 & 4.44 & 2.54 & 5.61 & 1.00 & 4.00 & 2.11 & 4.92 & 6 \\
Qwen-Image-Edit-2512           & 5.33 & 1.94 & 4.89 & 1.76 & 5.22 & 2.10 & 5.11 & 2.45 & 5.14 & 5 \\
FireRed-Image-Edit-1.1         & 5.20 & 1.94 & 4.47 & 1.75 & 4.40 & 1.58 & 4.00 & 1.93 & 4.52 & 7 \\
\rowcolor{gray!15}
Med-Banana\,(Qwen)             & 6.31 & 1.36 & 5.31 & 1.31 & 5.06 & 1.68 & 4.69 & 1.83 & 5.34 & 4 \\
\rowcolor{gray!15}
Med-Banana\,(FireRed)          & 5.84 & 1.69 & 5.26 & 1.55 & 5.26 & 1.48 & 5.05 & 2.06 & 5.36 & 3 \\

\bottomrule
\end{tabular}
}
\caption{\textbf{Main Experiment -- Medical image editing quality via multi-judge evaluation.}
We report Instruction Compliance, Pathological Plausibility, Anatomical Consistency, Imaging Bio-Fidelity, and their Overall average under automatic and expert judges. \textbf{Bold} indicates the best result, and \underline{underline} denotes the second-best.}
\label{tab:main_results}
\end{table*}

\noindent\textbf{Refiner tuning.}
Failed trajectories provide corrective supervision for prompt revision. For each rejected attempt, the refiner observes the failed edited image, previous instructions, the refinement history, and the current verifier reason, then learns to revise positive and negative prompt for the next attempt. This teaches the refiner to translate concrete failure evidence into an improved editing instruction.

\subsection{Edit--Verify--Refine Loop}
At inference time, Med-Banana starts from $p_0$ and repeatedly applies the same loop. At round $r$, the editor generates a candidate from the original source image and the current prompt pair $p_r$. The verifier evaluates the current edited image and returns a pass/fail decision with a reason. If the candidate is rejected, the refiner uses the previous edited image, prompt pair, refinement history, and verifier reason to produce the next prompt pair $p_{r+1}$.

The retry always edits from the original source image rather than from the failed candidate. This design prevents artifacts from accumulating across failed attempts and makes each round a controlled counterfactual attempt. Unless otherwise stated, the main system uses a refinement budget of $R=10$. The loop stops when the verifier accepts a candidate; if no candidate is accepted within the budget, the case is treated as unsuccessful and the saved intermediate attempts remain failed trajectories. This budget is evaluated explicitly in the ablation study, where we vary the number of refinement rounds and show that insufficient reflection under-corrects failures, while excessive reflection can drift from the target pathology.

\section{Experiments}
In this section, we evaluate whether success-and-failure trajectory supervision improves medical image editing. We compare state-of-the-art text-guided editors with Med-Banana editors on the same medical image editing benchmark, then evaluate visual preservation and distributional separability with objective probes.

\subsection{Experimental Setup}
\noindent\textbf{Evaluation set.}
The main experiment randomly selects 1000 editing tasks in Med-Banana-80K. Each example contains a healthy source image, a target disease, and the initial instruction \textit{add \{disease\} to this image}. The set covers brain MRI, fundus photography, and chest X-ray, with nine target disease categories: glioma, meningioma, pituitary tumor, cataract, AMD, diabetic retinopathy, cardiomegaly, pneumothorax, and pneumonia.

\noindent\textbf{State-of-the-Art Baselines.}
We benchmark Med-Banana with state-of-the-art proprietary and open-source image editors: GPT-Image-2~\cite{gpt_image_2}, Nano-Banana 1 \& 2~\cite{nano-banana}, Qwen-Image-Edit-2512~\cite{qwenimage}, FireRed-Image-Edit-1.1~\cite{firered}.

\noindent\textbf{Med-Banana configuration.}
The two Med-Banana variants use LoRA-adapted Qwen-Image-Edit-2511 and FireRed-Image-Edit-1.1 as editors. The verifier and refiner are separately finetuned Qwen3.5-9B vision-language models: the verifier scores candidate edits and returns a structured decision, while the refiner rewrites the prompt and negative prompt after a rejection. Med-Banana uses refinement budget $R=10$. Detailed settings are reported in Appendix~\ref{appendix:settings}.

\noindent\textbf{MLLM-as-Judge.}
We evaluate each edited candidate with two SOTA MLLMs: MiMo-V2.5~\cite{mimo}, a general-purpose model, and MedGemma-27B~\cite{medgemma}, a medically optimized model. Each judge compares the source and edited images with the task metadata, then assigns four 0--10 scores: target instruction compliance, biological/pathological plausibility, modality/anatomical consistency, and imaging-physics counterfactual fidelity. Rubric and prompt details are provided in Appendix~\ref{appendix:judge_prompt}.

\noindent\textbf{Expert-as-Judge.}
MLLM judges may have stylistic or model-specific biases~\cite{mllm-bias-1,mllm-bias-2}; therefore, we additionally conduct a single-blind expert evaluation over all seven medical image editors. Experts review the source image, edited candidate, target disease, modality, instruction, and a disease-specific imaging hint to evaluate the quality of medical image editing. Two medical annotation experts are trained to conduct quality evaluation using the same standards as the MLLM protocol. Full annotation settings and the interface are reported in Appendix~\ref{appendix:human_annotation}.

\begin{table*}[ht]
\centering
\large
\resizebox{0.8\linewidth}{!}{%
\setlength{\tabcolsep}{6pt}
\begin{tabular}{l|cc|cc|cc|c}
\toprule
\multirow{3}{*}{\textbf{Model}}
& \multicolumn{4}{c|}{\textbf{Similarity}}
& \multicolumn{2}{c|}{\textbf{Separability}}
& \textbf{Overall} \\
\cmidrule(lr){2-5}
\cmidrule(lr){6-7}
\cmidrule(lr){8-8}
& \multicolumn{2}{c|}{\textbf{LPIPS}$\downarrow$}
& \multicolumn{2}{c|}{\textbf{PSNR}$\uparrow$}
& \textbf{ResNet-18}$\downarrow$
& \textbf{ResNet-34}$\downarrow$
& \multirow[c]{2}{*}{\makecell[c]{\textbf{Average}\\\textbf{Rank}$\downarrow$}} \\
\cmidrule(lr){2-3}
\cmidrule(lr){4-5}
\cmidrule(lr){6-7}
& \textbf{Mean} & \textbf{Std.}
& \textbf{Mean} & \textbf{Std.}
& \textbf{Test Acc. (\%)} & \textbf{Test Acc. (\%)}
& \\
\midrule
GPT-Image-2                    & 0.229 & 0.117 & 21.64 & 4.01 & 63.00 & \underline{65.00} & 4.00 \\
Nano-Banana-2                  & 0.275 & 0.130 & 18.93 & 4.55 & 73.00 & \underline{65.00} & 5.25 \\
Nano-Banana                    & 0.250 & 0.112 & 18.72 & 6.41 & \textbf{55.00} & 70.00 & 4.75 \\
Qwen-Image-Edit-2512           & \textbf{0.127} & 0.082 & \underline{24.05} & 4.52 & 64.00 & 71.00 & 3.50 \\
FireRed-Image-Edit-1.1         & 0.166 & 0.086 & 22.27 & 4.69 & 82.00 & 79.00 & 5.50 \\
\rowcolor{gray!15}
Med-Banana\,(Qwen)             & 0.158 & 0.101 & 22.69 & 5.43 & 62.00 & \underline{65.00} & \underline{2.75} \\
\rowcolor{gray!15}
Med-Banana\,(FireRed)          & \underline{0.146} & 0.107 & \textbf{24.57} & 5.39 & \underline{59.00} & \textbf{64.00} & \textbf{1.50} \\
\bottomrule
\end{tabular}
}
\caption{\textbf{Objective Metrics -- Real/Synthetic Separability and Source Preservation.}
Methods are evaluated using LPIPS, PSNR, and ResNet-18/34 real-vs-edited test accuracy. Average Rank summarizes overall objective performance. \textbf{Bold} indicates the best result, and \underline{underline} denotes the second-best.}
\label{tab:image_similarity}
\end{table*}

\noindent\textbf{Source Preservation.}
We complement judge-based evaluation with objective image similarity between each source image and its edited output. Following the implementation in the objective-metric pipeline, Table~\ref{tab:image_similarity} reports PSNR~\cite{PSNR} and LPIPS~\cite{LPIPS} metrics. PSNR measures pixel-level preservation between medical images, where higher values indicate closer source-image fidelity. LPIPS measures perceptual pixel feature distance, where lower values indicate closer perceptual similarity. These metrics quantify whether an editor preserves non-target image content and modality-specific texture. See detailed settings in Appendix~\ref{appendix:objective_metrics}.

\noindent\textbf{Real/Synthetic separability.}
Medical image synthesis studies commonly assess perceptual or distributional realism through real-versus-synthetic discrimination ~\cite{med-edit-gan,C2ST}. We therefore introduce a real/synthetic classifier. For each editor, we sample 100 real pathology images exhibiting the corresponding diseases and the 100 edited images edited toward the same target diseases. We split these samples into train and test sets with a 1:1 ratio and train ResNet-18 and ResNet-34 classifiers for 50 epochs to distinguish real target-disease images from targeted editing outputs. This setup follows the logic of a classifier two-sample test: if the edited images contain obvious global artifacts or distributional shortcuts, the classifier should be able to separate them from real target-disease images; test accuracy closer to random guessing indicates stronger real-synthetic indistinguishability. We report final-epoch test accuracy, where lower is better.

\subsection{Main Results}
Tables~\ref{tab:main_results} and~\ref{tab:image_similarity} show that Med-Banana consistently improves the two open base editors, Qwen-Image-Edit and FireRed-Image-Edit. The key comparison is therefore not only against proprietary systems, but also against each base editor before trajectory supervision and verification-refinement.

\paragraph{MLLM-as-Judge.}
Under MiMo-V2.5, both Med-Banana variants reach 7.93 average score and rank second overall. Relative to their base editors, Med-Banana improves Qwen-Image-Edit by 0.16 points (7.77 to 7.93) and FireRed-Image-Edit by 0.24 points (7.69 to 7.93). The largest gain is in \textit{Imaging Bio-Fidelity}: Med-Banana (Qwen) improves from 7.65 to 8.02, and Med-Banana (FireRed) improves from 7.44 to 8.11. This directly supports the role of trajectory supervision in medical image editing: the model learns to introduce the requested pathology while preserving modality texture, surrounding anatomy, and bio-fidelity.

The stricter MedGemma-27B judge shows the same base-editor improvement with larger margins. Med-Banana (Qwen) improves over Qwen-Image-Edit by 1.06 points in overall average (3.04 to 4.10), and Med-Banana (FireRed) improves over FireRed-Image-Edit by 1.27 points (2.70 to 3.97). Thus, the main contribution is a consistent lift over open medical image editing backbones, rather than a result tied to one judge or one rubric dimension.

\paragraph{Expert-as-Judge.}
The annotators' assessment confirms the same direction. Med-Banana (Qwen) improves over Qwen-Image-Edit by 0.20 points in overall judge score (5.14 to 5.34), and Med-Banana (FireRed) improves over FireRed-Image-Edit by 0.84 points (4.52 to 5.36). The expert judges still rank proprietary systems highly, but the controlled comparison against Qwen and FireRed shows that our method improves the medical editing ability of both open base editors.

The judge scores also reveal where medical image editing remains difficult. Among modalities, chest X-ray obtains the highest average score (6.36), followed by fundus photography (5.90), while brain MRI is substantially lower (4.79). At the disease level, pneumothorax, cardiomegaly, and AMD receive the strongest scores, whereas pituitary tumor is the most difficult category with an average score of 3.93. This pattern suggests that large structural or intensity changes in brain MRI are more likely to break anatomical consistency and imaging bio-fidelity, while chest X-ray and fundus edits are judged more stable under the current edit--verify--refine pipeline.

\subsection{Objective Metrics}
Table~\ref{tab:image_similarity} highlights the objective contribution of Med-Banana. Compared with FireRed-Image-Edit, Med-Banana (FireRed) improves PSNR by 2.30 dB (22.27 to 24.57), reduces LPIPS from 0.166 to 0.146, and lowers real-synthetic classifier accuracy by 26 and 17 percentage points for ResNet-18 and ResNet-34. Compared with Qwen-Image-Edit, Med-Banana (Qwen) reduces classifier accuracy by 14 and 25 percentage points, moving generated medical editing outputs closer to the real-image distribution under both probes. These gains give the two Med-Banana variants the best overall objective ranks, 2.00 and 2.25, ahead of their FireRed and Qwen base editors at 5.25 and 3.50.

These objective results are central to our claim. The improvement is not only a judge preference for more visible disease patterns; the generated outputs also become harder to separate from real medical images and, for the FireRed backbone, better preserve source-image fidelity. This shows that success-and-failure trajectory supervision contributes directly to the editing capability required in medical images: localized pathology insertion without broad distributional artifacts.

\begin{table}[t]
\centering
\resizebox{\columnwidth}{!}{%
\setlength{\tabcolsep}{5pt}
\renewcommand{\arraystretch}{1.15}
\begin{tabular}{l|ccccc}
\toprule
\textbf{Variant}
& \textbf{Target}
& \textbf{Plaus.}
& \textbf{Anatomy}
& \textbf{Fidelity}
& \makecell[c]{\textbf{Overall}\\\textbf{Average}} \\
\midrule

\rowcolor{blue!15}
\multicolumn{6}{l}{\textbf{Component Tuning}} \\
w/o verifier       & 4.03 & 2.51 & 5.64 & 3.15 & 3.83 \\
w/o refiner        & 3.38 & 2.33 & 5.94 & 3.48 & 3.78 \\
w/o editor         & 4.23 & 2.67 & 5.62 & 2.95 & 3.87 \\

\midrule
\rowcolor{blue!15}
\multicolumn{6}{l}{\textbf{Refine Budget}} \\
$R=0$                      & 3.74 & 2.33 & 4.15 & 2.34 & 3.14 \\
$R=1$                      & 3.93 & 2.40 & 5.30 & 2.88 & 3.63 \\
$R=5$                      & 3.78 & 2.38 & 5.96 & 3.29 & 3.85 \\
 \textbf{$R=10$}                    &  \textbf{8.07} &  \textbf{7.06} &  \textbf{8.57} &  \textbf{8.02} & \textbf{7.93} \\
$R=20$                     & 3.99 & 2.51 & 5.80 & 3.15 & 3.86 \\

\bottomrule
\end{tabular}
}
\caption{\textbf{Ablation of trajectory-supervised editing.}
Component ablations remove verifier tuning, refiner tuning, or editor tuning, while reflection-budget ablations vary the number of refinement rounds $R$.}
\label{tab:ablation}
\end{table}

\section{Ablation Studies}

We conduct ablation studies to isolate how success-and-failure trajectory supervision contributes to medical image editing quality. As shown in Table~\ref{tab:ablation}, all variants are evaluated on the same 100 editing examples using MiMo-V2.5 as the MLLM-as-Judge. The judge reports four failure-taxonomy-derived dimensions: target instruction compliance (Target), biological and pathological plausibility (Plaus.), modality and anatomical consistency (Anatomy), and imaging-physics counterfactual fidelity (Fidelity). We use the overall average as the main summary metric.

\noindent\textbf{Contribution of Trajectory Supervision.}
The component ablations validate the role of success and failure supervision in the Med-Banana loop. Removing editor tuning (\textit{w/o editor}) removes the direct supervision from successful trajectories, reducing the overall average to 3.87. Removing the verifier or refiner removes two forms of failure supervision: candidate assessment from rejected edits and feedback-guided prompt revision from refinement traces. These variants reach only 3.83 and 3.78 overall average, respectively. In contrast, the full system with the default reflection budget achieves 7.93, with consistent gains across Target, Plausibility, Anatomy, and Fidelity. This shows that trajectory supervision does not merely improve a single visual attribute; it improves the balance between target disease insertion, medical plausibility, anatomical consistency, and counterfactual fidelity.

\noindent\textbf{Test-Time Reflection Scaling.}
We further study the refinement budget of the prompt-refinement loop. Here, $R$ denotes the maximum number of refinement rounds allowed at test time: the system returns once the verifier accepts a candidate, or stops after the budget is exhausted. Without reflection ($R=0$), the system lacks feedback correction and obtains the lowest overall score of 3.14. Increasing the budget to $R=1$ and $R=5$ improves anatomical consistency and fidelity, but the overall scores remain far below the full setting. The default budget $R=10$ performs best, reaching 7.93 overall and the highest score on every dimension. Pushing the loop to $R=20$ degrades the result to 3.86, indicating that excessive refinement can over-correct the prompt, drift away from the target pathology, or introduce visual artifacts. These results suggest that Med-Banana benefits from test-time reflection within a bounded refinement range.

Overall, supervised trajectory learning provides a stable refinement policy, while test-time reflection supplies controlled feedback at inference. Their combination is essential for the quality-controlled editing behavior of Med-Banana.

\begin{figure}[t]
    \centering
    \includegraphics[width=\linewidth]{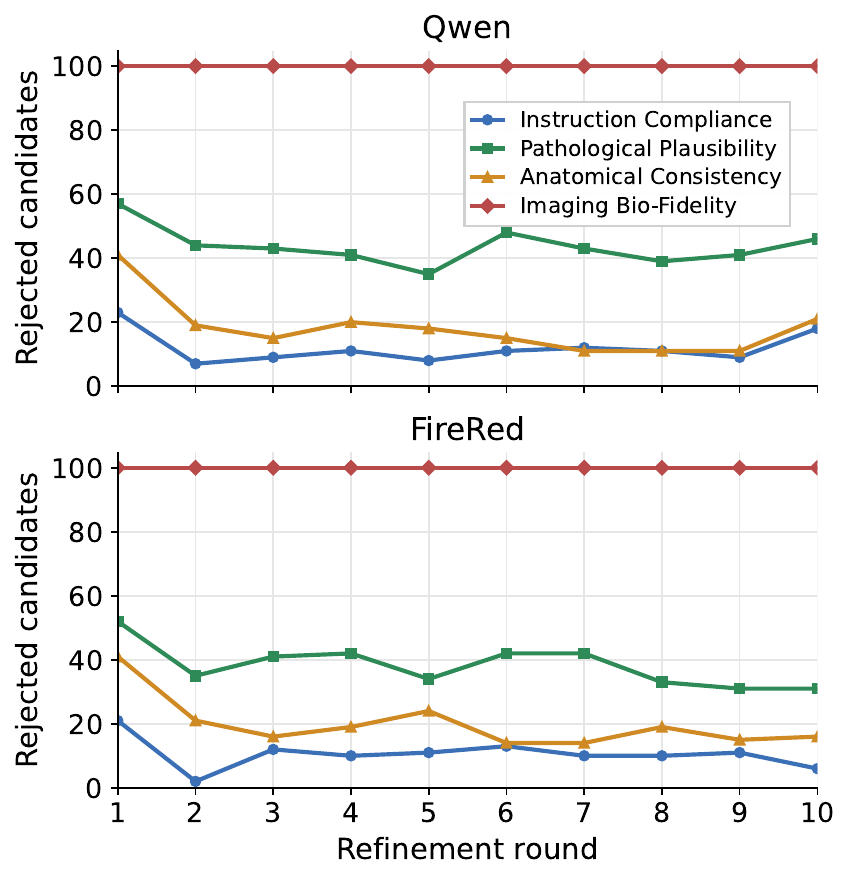}
    \caption{\textbf{Failure reasons over refinement rounds.}
    We count rejected candidates from rounds 1--10 and categorize each verifier reason into four diagnostic dimensions. Both Qwen and FireRed are dominated by imaging bio-fidelity failures, while instruction compliance errors are much less frequent after the first round.}
    \label{fig:trajectory_dimensions}
\end{figure}
\begin{figure}[t]
    \centering
    \includegraphics[width=\linewidth]{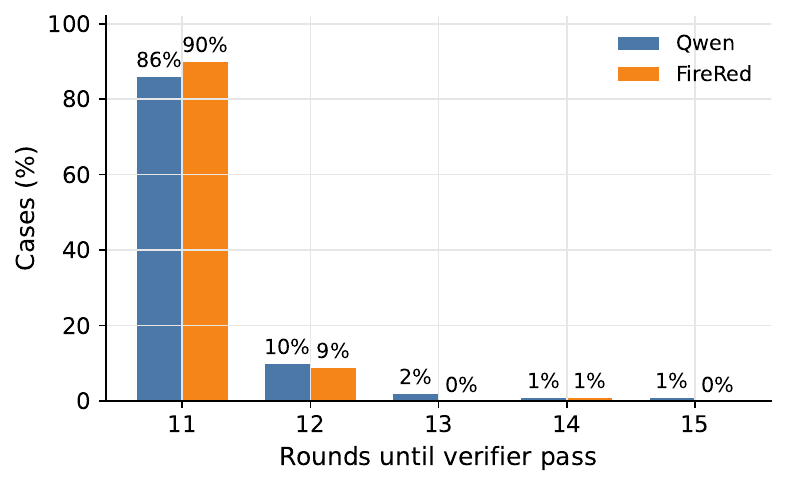}
    \caption{\textbf{Rounds required before verifier acceptance.}
    Bars report the percentage of editing cases that first pass the verifier at each round under the default refinement budget $R=10$. Most cases require multiple feedback rounds before acceptance, while a small tail remains difficult within the budget.}
    \label{fig:trajectory_rounds}
\end{figure}

\section{Refinement Trajectory Analysis}
\label{sec:trajectory_analysis}
We analyze instruction-refinement traces before verifier acceptance to characterize the supervision exposed by the edit--verify--refine loop. The analysis covers 100 cases for each editor variant, yielding 1,021 rejected candidates for Qwen and 1,012 for FireRed before the first accepted edit is reached.

\paragraph{Rejections are dominated by imaging-fidelity failures.}
Figure~\ref{fig:trajectory_dimensions} shows that imaging bio-fidelity is the dominant rejection dimension for both editors, appearing in every rejected candidate. The remaining dimensions are substantially less frequent: pathological plausibility appears in 452 Qwen and 388 FireRed rejections, anatomical consistency in 192 and 206, and instruction compliance in 135 and 112. The same hierarchy persists late in the loop. At round 10, all 100 rejected candidates from each editor still fail imaging bio-fidelity, while Qwen has 46 pathological-plausibility, 21 anatomical-consistency, and 18 instruction-compliance failures; FireRed has 31, 16, and 6. These trends indicate that refinement often corrects missing or misplaced disease evidence before it resolves image-formation defects, including local texture mismatch, boundary artifacts, contrast inconsistency, and modality-specific artifacts.

\paragraph{Acceptance concentrates after multiple rejected attempts.}
Figure~\ref{fig:trajectory_rounds} shows that verifier acceptance is concentrated after repeated rejection. Under the default $R=10$ setting, 86\% of Qwen cases and 90\% of FireRed cases first pass at round 11, with the remaining cases forming a small tail up to rounds 15 and 14. Thus, medical editing quality is rarely recovered by a single revision. The loop is better understood as conservative candidate selection under disease-specific and image-fidelity constraints, rather than smooth monotonic optimization. Its rejected rounds define the supervision boundary between visually plausible but invalid edits and verifier-accepted medical edits.

\section{Conclusion}
In this work, we presented \textbf{Med-Banana}, a framework for learning quality-controlled medical image editing from success-and-failure trajectories. We introduced \textbf{Med-Banana-80K}, a large-scale trajectory resource that records accepted edits, rejected candidates, verifier rationales, and prompt refinements, exposing the supervision that final-edit datasets typically discard. Building on it, Med-Banana couples an editor, verifier, and refiner in an edit--verify--refine loop, converting accepted attempts into positive editing supervision and rejected attempts into diagnostic and corrective supervision. Across MLLM-based evaluation, blind expert assessment, ablations, and objective source-preservation and real--synthetic separability probes, Med-Banana consistently improves open medical image editors. This work reframes failed medical edits as reusable training signal, offering a trajectory-level foundation for more controllable and reliable medical image editing.

\section*{Limitations}
We discuss three main limitations of this work.
First, Med-Banana-80K currently focuses on three representative 2D imaging modalities: chest X-ray, brain MRI, and fundus photography. Although the framework is not tied to specific modalities, extending it to additional settings such as CT, ultrasound, pathology, and volumetric imaging would improve its clinical coverage. Second, our evaluation combines MLLM judges, blind expert assessment, and objective probes, but the scale of expert evaluation remains limited. Future work could include larger multi-expert studies and broader external validation to further assess generalization across institutions, devices, and disease distributions. Third, Med-Banana is designed for research on medical image editing and quality control rather than direct clinical use. Before any clinical-facing deployment, the system would require expert validation, provenance tracking, and task-specific safety review.



\bibliography{custom}

@article{prism,
  title={Prism: High-resolution \& precise counterfactual medical image generation using language-guided stable diffusion},
  author={Kumar, Amar and Kriz, Anita and Havaei, Mohammad and Arbel, Tal},
  journal={arXiv preprint arXiv:2503.00196},
  year={2025}
}

@inproceedings{radedit,
  title={Radedit: stress-testing biomedical vision models via diffusion image editing},
  author={P{\'e}rez-Garc{\'\i}a, Fernando and Bond-Taylor, Sam and Sanchez, Pedro P and van Breugel, Boris and Castro, Daniel C and Sharma, Harshita and Salvatelli, Valentina and Wetscherek, Maria TA and Richardson, Hannah and Lungren, Matthew P and others},
  booktitle={European Conference on Computer Vision},
  pages={358--376},
  year={2024},
  organization={Springer}
}

@inproceedings{mededit,
  title={Mededit: Counterfactual diffusion-based image editing on brain mri},
  author={Alaya, Malek Ben and Lang, Daniel M and Wiestler, Benedikt and Schnabel, Julia A and Bercea, Cosmin I},
  booktitle={International Workshop on Simulation and Synthesis in Medical Imaging},
  pages={167--176},
  year={2024},
  organization={Springer}
}

@article{medgen-bench,
  title={MedGEN-Bench: Contextually entangled benchmark for open-ended multimodal medical generation},
  author={Yang, Junjie and Yan, Yuhao and Wu, Gang and Wang, Yuxuan and Liang, Ruoyu and Jiang, Xinjie and Wan, Xiang and Fan, Fenglei and Zhang, Yongquan and Qin, Feiwei and others},
  journal={arXiv preprint arXiv:2511.13135},
  year={2025}
}

@article{miedb-bench,
  title={MieDB-100k: A Comprehensive Dataset for Medical Image Editing},
  author={Lai, Yongfan and Qian, Wen and Liu, Bo and Li, Hongyan and Luo, Hao and Wang, Fan and Zhuang, Bohan and Hong, Shenda},
  journal={arXiv preprint arXiv:2602.09587},
  year={2026}
}

@inproceedings{MedEBench,
  title={MedEBench: Diagnosing Reliability in Text-Guided Medical Image Editing},
  author={Liu, Minghao and He, Zhitao and Fan, Zhiyuan and Wang, Qingyun and Fung, Yi R},
  booktitle={Findings of the Association for Computational Linguistics: EMNLP 2025},
  pages={767--791},
  year={2025}
}

@article{med-edit-gan,
  title={A multimodal comparison of latent denoising diffusion probabilistic models and generative adversarial networks for medical image synthesis},
  author={M{\"u}ller-Franzes, Gustav and Niehues, Jan Moritz and Khader, Firas and Arasteh, Soroosh Tayebi and Haarburger, Christoph and Kuhl, Christiane and Wang, Tianci and Han, Tianyu and Nolte, Teresa and Nebelung, Sven and others},
  journal={Scientific reports},
  volume={13},
  number={1},
  pages={12098},
  year={2023},
  publisher={Nature Publishing Group UK London}
}

@article{selfimprove_gen_medical,
  title={Self-improving generative foundation model for synthetic medical image generation and clinical applications},
  author={Wang, Jinzhuo and Wang, Kai and Yu, Yunfang and Lu, Yuxing and Xiao, Wenchao and Sun, Zhuo and Liu, Fei and Zou, Zixing and Gao, Yuanxu and Yang, Lei and others},
  journal={Nature Medicine},
  volume={31},
  number={2},
  pages={609--617},
  year={2025},
  publisher={Nature Publishing Group US New York}
}

@article{align_synth_med,
  title={Aligning synthetic medical images with clinical knowledge using human feedback},
  author={Sun, Shenghuan and Goldgof, Greg and Butte, Atul and Alaa, Ahmed M},
  journal={Advances in Neural Information Processing Systems},
  volume={36},
  pages={13408--13428},
  year={2023}
}

@article{medforge,
  title={MedForge: Interpretable Medical Deepfake Detection via Forgery-aware Reasoning},
  author={Chen, Zhihui and He, Kai and Lei, Qingyuan and Pu, Bin and Zhang, Jian and Xu, Yuling and Feng, Mengling},
  journal={arXiv preprint arXiv:2603.18577},
  year={2026}
}

@inproceedings{latent_brain,
  title={Latent drifting in diffusion models for counterfactual medical image synthesis},
  author={Yeganeh, Yousef and Farshad, Azade and Charisiadis, Ioannis and Hasny, Marta and Hartenberger, Martin and Ommer, Bj{\"o}rn and Navab, Nassir and Adeli, Ehsan},
  booktitle={Proceedings of the IEEE/CVF Conference on Computer Vision and Pattern Recognition},
  pages={7685--7695},
  year={2025}
}

@inproceedings{generation_fundus,
  title={Generation of structurally realistic retinal fundus images with diffusion models},
  author={Go, Sojung and Ji, Younghoon and Park, Sang Jun and Lee, Soochahn},
  booktitle={Proceedings of the IEEE/CVF conference on computer vision and pattern recognition},
  pages={2335--2344},
  year={2024}
}

@inproceedings{LPIPS,
  title={The unreasonable effectiveness of deep features as a perceptual metric},
  author={Zhang, Richard and Isola, Phillip and Efros, Alexei A and Shechtman, Eli and Wang, Oliver},
  booktitle={Proceedings of the IEEE conference on computer vision and pattern recognition},
  pages={586--595},
  year={2018}
}

@article{PSNR,
  title={Similarity and quality metrics for MR image-to-image translation},
  author={Dohmen, Melanie and Klemens, Mark A and Baltruschat, Ivo M and Truong, Tuan and Lenga, Matthias},
  journal={Scientific Reports},
  volume={15},
  number={1},
  pages={3853},
  year={2025},
  publisher={Nature Publishing Group UK London}
}

@article{RefineEdit,
  title={An LLM-LVLM Driven Agent for Iterative and Fine-Grained Image Editing},
  author={Liang, Zihan and Sun, Jiahao and Ma, Haoran},
  journal={arXiv preprint arXiv:2508.17435},
  year={2025}
}

@article{quality-image-edit,
  title={CAMEO: A Conditional and Quality-Aware Multi-Agent Image Editing Orchestrator},
  author={Pu, Yuhan and Zheng, Hao and Mo, Ziqian and Zhang, Hill and Fan, Tianyi and Wu, Shuhong and Wei, Jiaheng},
  journal={arXiv preprint arXiv:2604.03156},
  year={2026}
}

@article{image-edit-r1,
  title={ImageEdit-R1: Boosting Multi-Agent Image Editing via Reinforcement Learning},
  author={Zhao, Yiran and Ye, Yaoqi and Liu, Xiang and Shieh, Michael Qizhe and Bui, Trung},
  journal={arXiv preprint arXiv:2603.08059},
  year={2026}
}

@inproceedings{mllm-bias-1,
  title={Mllm-as-a-judge: Assessing multimodal llm-as-a-judge with vision-language benchmark},
  author={Chen, Dongping and Chen, Ruoxi and Zhang, Shilin and Wang, Yaochen and Liu, Yinuo and Zhou, Huichi and Zhang, Qihui and Wan, Yao and Zhou, Pan and Sun, Lichao},
  booktitle={Forty-first International Conference on Machine Learning},
  year={2024}
}

@inproceedings{mllm-bias-2,
  title={Judge anything: Mllm as a judge across any modality},
  author={Pu, Shu and Wang, Yaochen and Chen, Dongping and Chen, Yuhang and Wang, Guohao and Qin, Qi and Zhang, Zhongyi and Zhang, Zhiyuan and Zhou, Zetong and Gong, Shuang and others},
  booktitle={Proceedings of the 31st ACM SIGKDD Conference on Knowledge Discovery and Data Mining V. 2},
  pages={5742--5753},
  year={2025}
}

@article{C2ST,
  title={Revisiting classifier two-sample tests},
  author={Lopez-Paz, David and Oquab, Maxime},
  journal={arXiv preprint arXiv:1610.06545},
  year={2016}
}

@article{mimo,
  title={Mimo-v2-flash technical report},
  author={Xiao, Bangjun and Xia, Bingquan and Yang, Bo and Gao, Bofei and Shen, Bowen and Zhang, Chen and He, Chenhong and Lou, Chiheng and Luo, Fuli and Wang, Gang and others},
  journal={arXiv preprint arXiv:2601.02780},
  year={2026}
}

@article{medgemma,
  title={Medgemma technical report},
  author={Sellergren, Andrew and Kazemzadeh, Sahar and Jaroensri, Tiam and Kiraly, Atilla and Traverse, Madeleine and Kohlberger, Timo and Xu, Shawn and Jamil, Fayaz and Hughes, C{\'\i}an and Lau, Charles and others},
  journal={arXiv preprint arXiv:2507.05201},
  year={2025}
}

@article{qwenimage,
  title={Qwen-image technical report},
  author={Wu, Chenfei and Li, Jiahao and Zhou, Jingren and Lin, Junyang and Gao, Kaiyuan and Yan, Kun and Yin, Sheng-ming and Bai, Shuai and Xu, Xiao and Chen, Yilei and others},
  journal={arXiv preprint arXiv:2508.02324},
  year={2025}
}

@article{firered,
  title={Firered-image-edit-1.0 technical report},
  author={Team, Super Intelligence and Qiao, Changhao and Hui, Chao and Li, Chen and Wang, Cunzheng and Song, Dejia and Zhang, Jiale and Li, Jing and Xiang, Qiang and Wang, Runqi and others},
  journal={arXiv preprint arXiv:2602.13344},
  year={2026}
}

@article{gpt_image_2,
  title={Openai gpt-5 system card},
  author={Singh, Aaditya and Fry, Adam and Perelman, Adam and Tart, Adam and Ganesh, Adi and El-Kishky, Ahmed and McLaughlin, Aidan and Low, Aiden and Ostrow, AJ and Ananthram, Akhila and others},
  journal={arXiv preprint arXiv:2601.03267},
  year={2025}
}

@article{nano-banana,
  title={Gemini 2.5: Pushing the frontier with advanced reasoning, multimodality, long context, and next generation agentic capabilities},
  author={Comanici, Gheorghe and Bieber, Eric and Schaekermann, Mike and Pasupat, Ice and Sachdeva, Noveen and Dhillon, Inderjit and Blistein, Marcel and Ram, Ori and Zhang, Dan and Rosen, Evan and others},
  journal={arXiv preprint arXiv:2507.06261},
  year={2025}
}

@inproceedings{odir,
  title={A benchmark of ocular disease intelligent recognition: One shot for multi-disease detection},
  author={Li, Ning and Li, Tao and Hu, Chunyu and Wang, Kai and Kang, Hong},
  booktitle={International symposium on benchmarking, measuring and optimization},
  pages={177--193},
  year={2020},
  organization={Springer}
}

@article{mimic,
  title={MIMIC-III, a freely accessible critical care database},
  author={Johnson, Alistair EW and Pollard, Tom J and Shen, Lu and Lehman, Li-wei H and Feng, Mengling and Ghassemi, Mohammad and Moody, Benjamin and Szolovits, Peter and Anthony Celi, Leo and Mark, Roger G},
  journal={Scientific data},
  volume={3},
  number={1},
  pages={1--9},
  year={2016},
  publisher={Nature Publishing Group}
}

@misc{mri,
	title={Brain Tumor MRI Dataset},
	url={https://www.kaggle.com/dsv/2645886},
	DOI={10.34740/KAGGLE/DSV/2645886},
	publisher={Kaggle},
	author={Msoud Nickparvar},
	year={2021}
}

\newpage
\appendix

\section{Med-Banana-80K Construction Details}
\label{appendix:construction}
Med-Banana-80K is constructed as an editing trajectory resource rather than a final-pair collection. Each source image is paired with a target disease and an initial editing instruction. The initial instruction is intentionally simple, using the template \textit{add \{disease\} to this image}. The editor generates a candidate image, and the verifier compares the source and candidate side by side. If the candidate is accepted, the record is stored as a successful trajectory. If it is rejected, the failure reason is stored and passed to the refiner to produce the next prompt.

\paragraph{Trajectory fields.}
Each trajectory stores the source image identifier, modality, disease label, task type, prompt, candidate edited image, outcome, rejection reason when present, and refinement trace when a retry is generated. Successful trajectories therefore provide source--instruction--edited-image supervision for editor adaptation, while failed trajectories provide assessor and refiner supervision.

\paragraph{Refinement protocol.}
The retry image is always generated from the original source image, not from the previously failed candidate. This keeps each attempt anchored to the same clinical image and prevents error accumulation across rounds. In the main system, the default refinement budget is $R=10$: after the initial prompt, the system stops when the verifier accepts a candidate or when the budget is exhausted. The ablation study evaluates alternative budgets $R\in\{0,1,5,10,20\}$.

\begin{tcolorbox}[colback=gray!10, colframe=gray!50, title=Initial Editing Prompt Template, breakable]
\small
\textbf{User prompt:} ``add \{disease\} to this image''

\noindent\textbf{Example:} ``add meningioma to this image''
\end{tcolorbox}

\begin{table*}[h]
\centering
\small
\setlength{\tabcolsep}{4pt}
\renewcommand{\arraystretch}{1.15}
\begin{tabular}{>{\raggedright\arraybackslash}p{0.24\linewidth} >{\raggedright\arraybackslash}p{0.68\linewidth}}
\toprule
\textbf{Item} & \textbf{Setting} \\
\midrule
\rowcolor{blue!15}
\multicolumn{2}{l}{\textbf{Editor LoRA Fine-tuning}} \\
Training subset & 10K Med-Banana trajectories \\
Editor backbones & Qwen-Image-Edit-2511; FireRed-Image-Edit-1.1 \\
Supervision & successful source--instruction--edited-image triples \\
Adaptation method & LoRA fine-tuning \\
Frozen components & Qwen-Image text encoder and VAE \\
Trainable editor module & DiT LoRA \\
LoRA target modules & attention: \texttt{to\_q}, \texttt{to\_k}, \texttt{to\_v}, \texttt{add\_q\_proj}, \texttt{add\_k\_proj}, \texttt{add\_v\_proj}; output: \texttt{to\_out.0}, \texttt{to\_add\_out}; MLP/modulation: \texttt{img\_mlp.net.2}, \texttt{img\_mod.1}, \texttt{txt\_mlp.net.2}, \texttt{txt\_mod.1} \\
LoRA rank & 32 \\
Learning rate & $1\times10^{-4}$ \\
Weight decay & 0.01 \\
Dataset repeat & 1 \\
Max pixels & 1,048,576 \\
Gradient accumulation & 1 \\
Gradient checkpointing & enabled \\
Final editor checkpoint & epoch-9 LoRA checkpoint \\
Qwen-specific conditioning & zero conditional timestep enabled for Qwen only \\
Editor inference & bf16, 20 denoising steps, CFG 4.0, seed 0, addition and removal tasks \\
\midrule
\rowcolor{blue!15}
\multicolumn{2}{l}{\textbf{Verifier and Refiner VLM SFT}} \\
Training subset & 10K Med-Banana trajectories \\
Backbones & Qwen3.5-9B prompt-verifier; Qwen3.5-9B prompt-refiner \\
Adaptation method & LoRA SFT with ms-swift on system/user/assistant VLM conversations \\
Verifier supervision & edited-candidate assessment examples; output fields \texttt{qualified}, \texttt{has\_disease}, \texttt{structure\_reasonable}, \texttt{looks\_realistic}, and \texttt{reason} \\
Refiner supervision & failed-attempt context and verifier feedback; output fields \texttt{prompt} and \texttt{negative\_prompt} \\
LoRA settings & rank 64, alpha 128, dropout 0.05, target modules \texttt{all-linear}, saved modules \texttt{embed\_tokens} and \texttt{lm\_head} \\
Trainable VLM parts & ViT and aligner unfrozen; gradient checkpointing and ViT gradient checkpointing enabled \\
Optimization & bf16, 1 epoch, learning rate $1\times10^{-4}$, weight decay 0.1, cosine schedule, warmup ratio 0.05 \\
Batching and length & per-device train/eval batch size 4, gradient accumulation 8, packing and padding-free training enabled, maximum length 2048 \\
Distributed training & 2 processes, FlashAttention, DeepSpeed ZeRO-2, 16 dataset workers and 16 dataloader workers \\
Checkpointing and logging & save every 200 steps, keep 3 checkpoints, evaluate every 400 steps, log every 6 steps \\
\bottomrule
\end{tabular}
\caption{\textbf{Component fine-tuning settings.} The editor settings are shared by the Qwen and FireRed Med-Banana editors; the verifier and refiner are separate Qwen3.5-9B VLM SFT models trained from trajectory-derived inquiry--response examples.}
\label{tab:editor_lora_settings}
\end{table*}

\section{Experiment Settings}
\label{appendix:settings}
\subsection{Component Fine-tuning}
We fine-tune the editor, verifier, and refiner from the same 10K subset of Med-Banana trajectories. The editor uses successful trajectory records as positive source--instruction--edited-image triples. The verifier and refiner use the trajectory interaction records as VLM supervised fine-tuning examples: the verifier learns to assess edited candidates with structured JSON decisions, while the refiner learns to convert a failed attempt, its previous prompt pair, and verifier feedback into a revised prompt and negative prompt. This shared subset keeps all three modules tied to the same medical editing distribution while assigning each module the supervision signal that matches its role in the edit--verify--refine loop.

\subsection{Objective Metrics Details}
\label{appendix:objective_metrics}
We use objective probes to measure source-image preservation and real--synthetic separability. These probes are not intended to judge clinical correctness; they quantify whether an editor preserves non-target image content and avoids global synthetic artifacts.

\paragraph{Similarity metrics.}
Let $\boldsymbol{x}, \hat{\boldsymbol{x}} \in [0,1]^{3\times H\times W}$ denote the RGB source image and edited output loaded by the objective-metric pipeline. PSNR is computed from the RGB mean squared error,
\[
\operatorname{MSE}(\boldsymbol{x},\hat{\boldsymbol{x}})
= \frac{1}{3HW}\sum_{c=1}^{3}\sum_{h=1}^{H}\sum_{w=1}^{W}
\left(\boldsymbol{x}_{c,h,w}-\hat{\boldsymbol{x}}_{c,h,w}\right)^2
\]
\[
\operatorname{PSNR}(\boldsymbol{x},\hat{\boldsymbol{x}})
= 20\log_{10}\left(\frac{1}{\sqrt{\operatorname{MSE}(\boldsymbol{x},\hat{\boldsymbol{x}})}}\right)
\]
because the implementation normalizes pixel values to $[0,1]$. Higher PSNR therefore indicates stronger pixel-level source preservation.
LPIPS is computed with the official AlexNet-backbone LPIPS model. Before evaluation, the RGB tensors are linearly mapped to $[-1,1]$ as $\tilde{\boldsymbol{x}}=2\boldsymbol{x}-1$ and $\tilde{\hat{\boldsymbol{x}}}=2\hat{\boldsymbol{x}}-1$. The metric compares normalized deep features across AlexNet layers:
\[
\begin{aligned}
\operatorname{LPIPS}(\boldsymbol{x},\hat{\boldsymbol{x}})
&= \sum_l \frac{1}{H_lW_l}\sum_{h,w}
\left\|\boldsymbol{w}_l \odot \Delta_l(h,w)\right\|_2^2, \\
\Delta_l(h,w)
&= \hat{\phi}_l(\tilde{\boldsymbol{x}})_{h,w}
- \hat{\phi}_l(\tilde{\hat{\boldsymbol{x}}})_{h,w}.
\end{aligned}
\]
Here, $\hat{\phi}_l$ denotes channel-normalized AlexNet features and $\boldsymbol{w}_l$ are the learned LPIPS layer weights. Lower LPIPS indicates smaller perceptual feature distance. For both metrics, we report the mean and standard deviation over 100 paired source and edited images for each editor.

\paragraph{Classifier probes.}
We train ResNet-18 and ResNet-34 as binary real-vs-edited probes for each editor independently. Source images are labeled as real and edited outputs as synthetic. Each probe uses 100 real and 100 edited images with a balanced 5:5 train/test split per class, image size $224\times224$, batch size 16, AdamW optimization, learning rate $1\times10^{-4}$, and random seed 82. We train the networks from scratch without pretrained weights and report final-epoch test accuracy. Lower test accuracy indicates weaker real--synthetic separability.

\subsection{Verifier and Refiner Tuning Templates}
The verifier and refiner are tuned as separate Qwen3.5-9B vision-language models. The verifier uses edited-candidate assessment examples, while the refiner uses failed-attempt context and verifier feedback examples. The templates below summarize the training interface; implementation-specific wording is omitted.

\begin{tcolorbox}[colback=gray!10, colframe=gray!50, title=Verifier Inquiry--Response Template, breakable]
\small
\textbf{Input image.} The edited candidate image.

\noindent\textbf{Inquiry.} Given the target disease and modality, decide whether the edited image is a qualified medical edit. Check whether the target finding is present, whether the anatomy or structure remains reasonable, and whether the image appears realistic.

\noindent\textbf{Response schema.} Return one JSON object with \texttt{qualified}, \texttt{has\_disease}, \texttt{structure\_reasonable}, \texttt{looks\_realistic}, and \texttt{reason}. The reason should briefly justify the pass or failure decision.
\end{tcolorbox}

\begin{tcolorbox}[colback=gray!10, colframe=gray!50, title=Refiner Inquiry--Response Template, breakable]
\small
\textbf{Input image.} The failed edited candidate image from the previous round.

\noindent\textbf{Inquiry.} Given the previous prompt pair and verifier feedback, write a revised prompt pair for the next editing attempt. The revised prompt should address the concrete failure evidence, preserve the source image outside the target edit, and avoid repeating the same failure mode.

\noindent\textbf{Response schema.} Return one JSON object with \texttt{prompt} and \texttt{negative\_prompt}. The prompt describes the intended local edit, while the negative prompt lists visual artifacts and unintended changes to avoid.
\end{tcolorbox}

\subsection{MLLM-as-Judge Details}
\label{appendix:judge_prompt}
The automatic judges compare the original source image and the edited candidate image. MedGemma-27B is served through a local OpenAI-compatible endpoint, while MiMo-V2.5 is accessed through an OpenAI-compatible API. The MiMo script imports the MedGemma prompt source, so both judges share the same scoring standard. The main MedGemma run uses temperature 0.5 and concurrency 32; the main MiMo run uses temperature 0.0, concurrency 8, API retries, and JSON response formatting.

\begin{tcolorbox}[colback=gray!10, colframe=gray!50, title=MLLM-as-Judge User Prompt Template, breakable]
\small
\textbf{Task.} Score Image 2 against Image 1 for this medical addition task.

\noindent\textbf{Image roles.} Image 1 is the original source image before editing. Image 2 is the edited candidate image to score.

\noindent\textbf{Task metadata.} The prompt includes \texttt{task\_type: addition}, \texttt{modality}, \texttt{target\_disease}, \texttt{instruction}, and \texttt{sample\_stem}.

\noindent\textbf{Scoring rules.} The judge scores each dimension from 0 to 10 and does not calculate an overall score. It must judge Image 2 relative to Image 1 and the requested disease, use the full score range, set quality flags before scoring, and keep flags, reasons, failure modes, and scores mutually consistent.

\noindent\textbf{Dimensions.} The four dimensions are  Instruction Compliance, Pathological Plausibility, Anatomical Consistency, and Imaging Fidelity.

\noindent\textbf{Output schema.} The judge returns exactly one JSON object containing \texttt{role\_check}, \texttt{quality\_flags}, \texttt{scores}, \texttt{dimension\_reasons}, and \texttt{failure\_modes}. Each reason is a short sentence and \texttt{failure\_modes} contains at most three short strings.
\end{tcolorbox}



\subsection{Expert-as-Judge Annotation}
\label{appendix:human_annotation}
We build the expert-as-judge annotation tool from the same editing evaluation package used for the main experiments. The annotation covers all evaluated models, 100 edited images each. The target diseases are glioma, meningioma, pituitary tumor, cardiomegaly, pneumonia, pneumothorax, AMD, cataract, and diabetic retinopathy.

\paragraph{Blind annotation protocol.}
The interface is single-blind with respect to model identity. For each case, the annotator sees the original source image and the edited image, while the hidden record keeps the true system name, sample stem, modality, target disease, and instruction. Cases are shuffled with random seed 82 before annotation.

\paragraph{Displayed context.}
To keep the judgment focused on the requested edit, the interface displays the original image, the edited image to be scored, the target disease, modality, and instruction. It also provides a short disease-specific imaging hint. For example, chest X-ray hints describe expected radiographic findings such as cardiomegaly, pneumonia, and pneumothorax; fundus hints describe AMD, cataract, and diabetic retinopathy appearances; and brain MRI hints describe glioma, meningioma, and pituitary tumor locations and morphology. The hint is used as annotation guidance only and does not reveal the model identity.

\paragraph{Scoring rubric.}
Annotators assign four independent scores from 0 to 10: Instruction Compliance, Pathological Plausibility, Anatomical Consistency, and Imaging Fidelity. These dimensions match the automatic MLLM-as-Judge rubric. In the annotation guide, scores 8--10 indicate a clear and medically plausible edit with only minor issues, 6--7 indicate an acceptable edit with moderate imperfections, 4--5 indicate a weak or borderline edit, and 0--3 indicate failure such as missing target pathology, wrong disease, anatomical impossibility, non-medical content, or severe artifact.

\begin{tcolorbox}[colback=gray!10, colframe=gray!50, title=Med-Banana Edit Expert Rating Guidelines, breakable]
\small

\textbf{Task Description.}
Rate only the edited image on the right. The original image on the left is used only for comparison. The edited image should add the target disease manifestation while preserving the non-lesion regions of the original image as much as possible.

\textbf{Scoring Range.}
Each dimension is scored from 0 to 10.

\begin{itemize}
    \item \textbf{8--10: Excellent.} The target is clear, the location and medical appearance are reasonable, with only minor issues.
    \item \textbf{6--7: Acceptable.} The target is recognizable, but there are moderate flaws, incomplete supporting signs, or insufficiently natural integration.
    \item \textbf{4--5: Marginally usable.} The target is weak, ambiguous, poorly localized, or of limited medical significance.
    \item \textbf{0--3: Failure.} The target is missing, the disease is incorrect, the anatomy is wrong, the image has been globally replaced, text or symbols are used instead of a lesion, or clearly non-medical content appears.
\end{itemize}

Do not default to a score of 5. If the target lesion is clearly visible, approximately well located, and has no obvious medical contradiction, the score should generally fall in the 6--8 range, with deductions based on specific defects.

\textbf{Four Rating Dimensions.}

\begin{enumerate}
    \item \textbf{Target Lesion / Instruction Fulfillment.}
    Assess whether the edited image completes the disease-addition task specified in the instruction.
    A high score means that the target disease is clearly visible, approximately well located, and consistent with the original task.
    A low score means that the target is absent, the disease is incorrect, the target is extremely ambiguous, or the edited content does not support the intended disease.
    Slightly fake texture or unnatural edges should not by itself push this dimension very low, as long as the target remains recognizable.

    \item \textbf{Biological and Pathological Plausibility.}
    Assess whether the disease appearance is consistent with medical knowledge and pathological logic.
    A high score means that the lesion morphology, extent, progression pattern, and related tissue response are basically reasonable.
    A low score means that there are clear biological contradictions, such as impossible lesion morphology, unreasonable disease stage, or completely mismatched associated structures.
    Missing some ideal secondary signs may justify deductions, but recognizable lesions should not be treated as failures by default.

    \item \textbf{Modality and Anatomical Consistency.}
    Assess whether the edit preserves the corresponding imaging modality and major anatomical structures.
    A high score means that modality features such as brain MRI, fundus photography, or chest X-ray are well preserved, and lesion location and spatial relationships are reasonable.
    A low score means that key anatomical structures are disrupted, the lesion is placed in an impossible location, or the modality characteristics are clearly incorrect.
    Texture, noise, and blending issues should mainly be penalized in the fourth dimension, rather than being severely penalized again here.

    \item \textbf{Imaging Realism and Counterfactual Fidelity.}
    Assess whether the edited image looks like a real medical image and whether only the lesion region that should change has been altered.
    A high score means that noise, texture, edges, contrast, and scanner or device style look natural, and non-lesion regions are basically preserved.
    A low score means that there are obvious pasted artifacts, hard edges, inconsistent noise, local smearing, repeated textures, irrelevant global changes, or image corruption.
    If the target lesion remains clear and lies within a medical image, mild to moderate synthetic artifacts usually merit a score of 5--6, not automatically 0--4.
\end{enumerate}

\textbf{Scoring Principles.}
The four dimensions should be scored independently. Do not apply extreme repeated penalties across all dimensions for the same defect.
First determine whether the target disease is recognizable, then assess medical plausibility, anatomical consistency, and imaging realism.
Synthetic artifacts should mainly affect the imaging realism and counterfactual fidelity score, unless they make the target unrecognizable or anatomically impossible.



\end{tcolorbox}

\newpage

\section{Failure Mode Analysis and Qualitative Examples}
\label{appendix:failure_examples}

This appendix provides qualitative audits of rejected candidates from the default $R=10$ edit--verify--refine setting. The purpose is not to introduce another aggregate metric, but to make the stored trajectories inspectable: each example shows what the verifier rejects, which failure mode is exposed, and how later attempts differ before a verifier-accepted candidate is obtained. We use editing traces from the same run analysed in the main paper. The displayed rounds are illustrative checkpoints selected to cover recurring failure types, including image-formation artifacts, target absence, prompt drift, and modality-specific preservation errors.

\begin{figure*}[h]
    \centering
    \includegraphics[width=\linewidth]{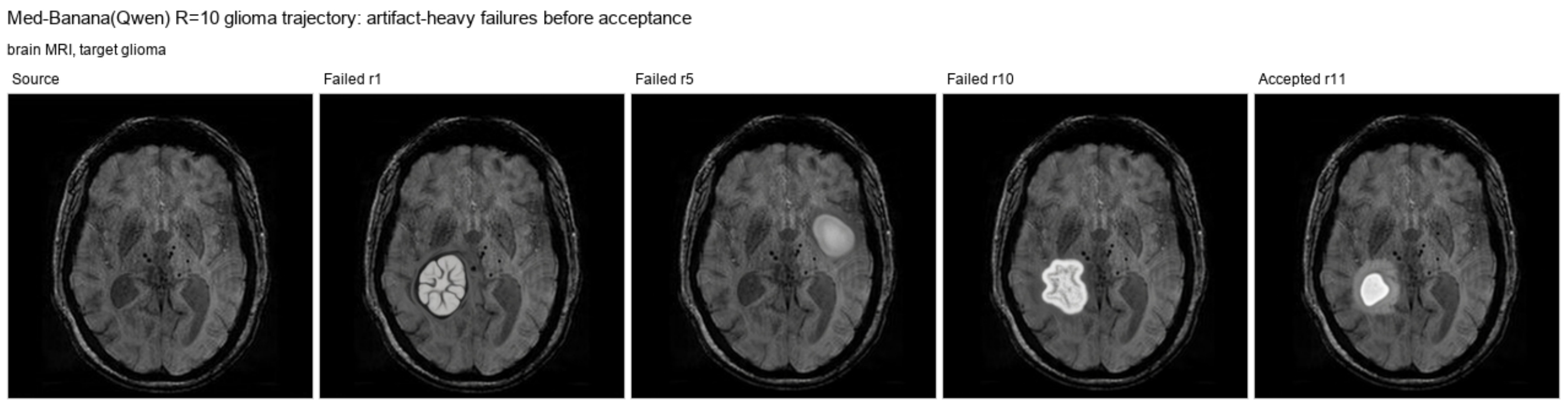}
    \caption{\textbf{Failure-to-acceptance trajectory for a Qwen glioma edit.}
    Early candidates introduce a visible lesion-like region, but the verifier rejects them due to pasted boundaries, unnatural internal texture, halo artifacts, and overly smooth lesion margins. Later attempts reduce some image-formation artifacts while preserving the intended abnormality. The final verifier-accepted candidate better matches the surrounding MRI texture while retaining glioma-like evidence.}
    \label{fig:appendix_qwen_glioma_trajectory}
\end{figure*}

\begin{figure*}[h]
    \centering
    \includegraphics[width=\linewidth]{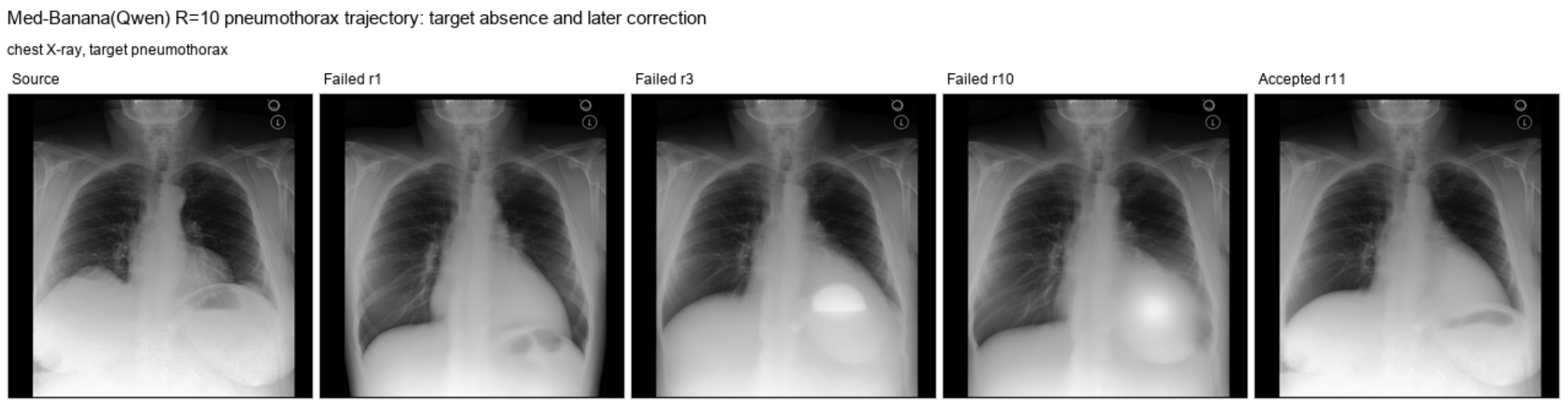}
    \caption{\textbf{Target-missing behavior in a Qwen pneumothorax trajectory.}
    Early candidates preserve the global chest X-ray structure but do not show a convincing pneumothorax pattern: lung markings remain visible near the chest wall or the generated change resembles an unrelated opacity. The final verifier-accepted candidate better expresses the intended radiographic change while keeping the overall X-ray layout stable.}
    \label{fig:appendix_pneumothorax_target_missing}
\end{figure*}

\begin{figure*}[h]
    \centering
    \includegraphics[width=\linewidth]{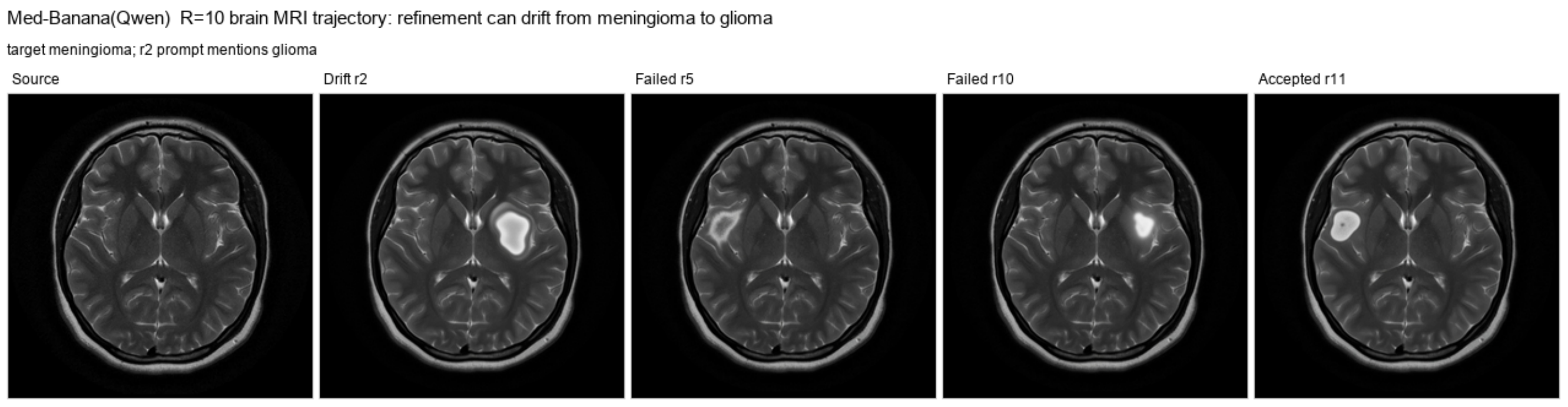}
    \caption{\textbf{Prompt drift in a brain MRI refinement trajectory.}
    The target disease is meningioma, but an intermediate refined prompt introduces glioma-like evidence. The trajectory therefore exposes two coupled errors: visual artifacts in the generated image and pathology-level drift in the refinement instruction.}
    \label{fig:appendix_prompt_drift}
\end{figure*}

\begin{figure*}[h]
    \centering
    \includegraphics[width=\linewidth]{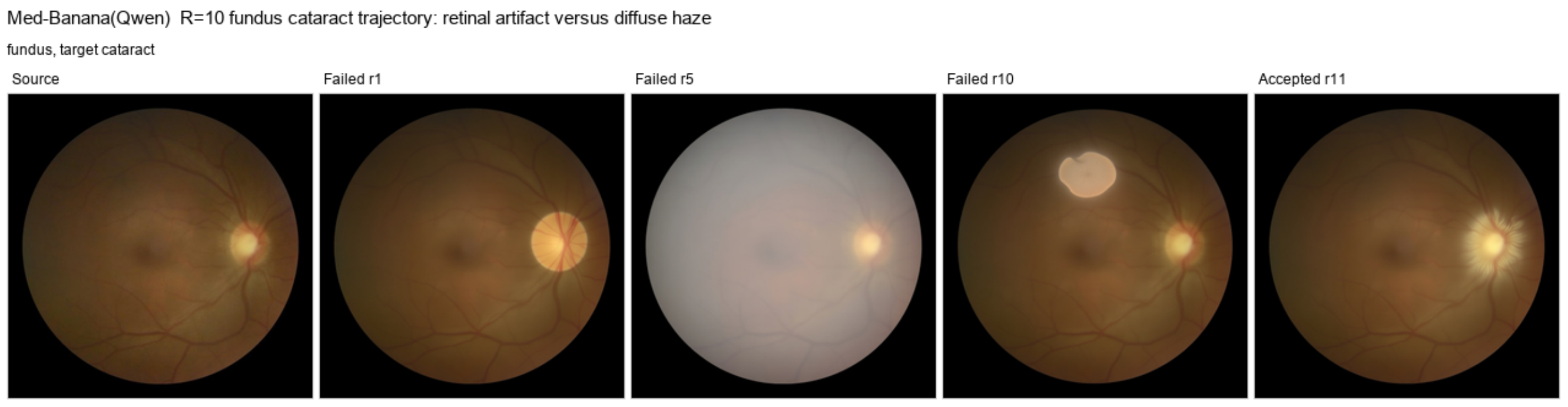}
    \caption{\textbf{Failure-to-acceptance trajectory for a Qwen fundus cataract edit.}
    Rejected candidates expose a modality-specific error pattern. One round over-whitens the entire fundus view, while another inserts a discrete retinal lesion. The final verifier-accepted candidate keeps the cataract signal closer to the expected image-formation mechanism: reduced retinal clarity without introducing a focal retinal mass.}
    \label{fig:appendix_fundus_failures}
\end{figure*}

\begin{figure}[h]
    \centering
    \includegraphics[width=\linewidth]{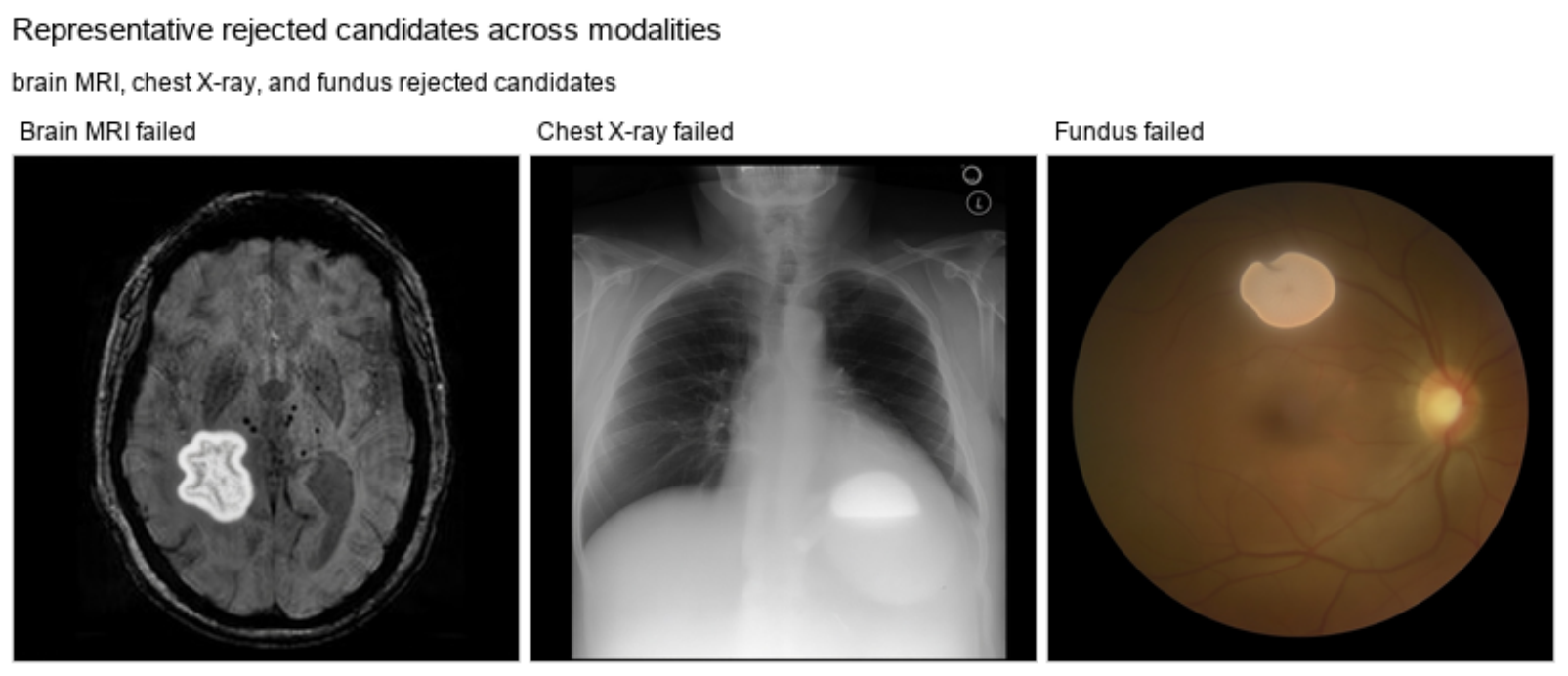}
    \caption{\textbf{Representative failure-to-acceptance examples across modalities.}
    The panel summarizes trajectory examples from brain MRI, chest X-ray, and fundus photography. Across modalities, the rejected candidates expose different failure mechanisms before a verifier-accepted candidate is obtained: lesion-boundary artifacts in MRI, insufficient or misplaced radiographic evidence in chest X-ray, and retinal texture or illumination artifacts in fundus imaging.}
    \label{fig:appendix_multimodal_trajectories}
\end{figure}

\paragraph{Image-formation artifacts are distinct from target absence.}
Figure~\ref{fig:appendix_qwen_glioma_trajectory} shows that the editor can attempt the requested disease from the first round while still producing an invalid medical edit. The rejected candidates are not failures because the target is entirely missing; they are failures because the inserted evidence is not integrated into the acquisition style of the source scan. Hard lesion boundaries, halo-like transitions, repeated internal texture, and local contrast mismatch make the abnormality appear digitally inserted rather than image-formed. This distinction is important for trajectory supervision: a final-pair dataset would only retain the verifier-accepted image, whereas the rejected candidates specify which visual cues make an otherwise target-visible edit unacceptable under the imaging-fidelity criterion.

\paragraph{Instruction compliance can fail even when image realism is preserved.}
Figure~\ref{fig:appendix_pneumothorax_target_missing} illustrates a complementary failure mode. Some candidates remain plausible as chest radiographs, but they do not encode the requested pathology. This case separates generic image realism from instruction compliance: preserving the thoracic layout is insufficient if the counterfactual sign is absent, misplaced, or confused with another opacity. The corresponding trajectory is useful because it provides a negative example where the refiner should preserve global structure while increasing disease-specific evidence. In this sense, the failed rounds supervise a more precise correction than simply asking the editor to make the image ``more realistic.''

\paragraph{Refinement traces expose prompt-level failure modes.}
Figure~\ref{fig:appendix_prompt_drift} shows a failure type that is difficult to detect from final edited images alone. The refined prompt can become overly specific and shift from the requested disease toward a related but different pathology. This is not merely an image-generation error; it is a refinement error in which feedback is converted into a new instruction that partially changes the target concept. Storing the rejected image together with the refined prompt makes this failure auditable. It also identifies a concrete constraint for future refiners: feedback should correct visual defects and missing signs while preserving the original disease target.

\paragraph{Fundus failures depend on disease-specific image formation.}
Figure~\ref{fig:appendix_fundus_failures} shows that fundus errors are not simply a matter of edit strength. A cataract edit should reduce retinal visibility through blur, haze, or contrast loss caused by media opacity. By contrast, a bright focal patch on the retina suggests a different retinal abnormality, and excessive whitening removes too much of the fundus acquisition structure. The trajectory therefore distinguishes between a weak cataract signal, an over-applied global degradation, and a wrong-pathology artifact. This example illustrates why the verifier must assess the mechanism by which the target disease appears, rather than only whether a visible abnormality has been added.

\paragraph{Failure modes vary systematically across modalities.}
Figure~\ref{fig:appendix_multimodal_trajectories} shows that rejected candidates differ substantially across imaging modalities. Brain MRI edits are sensitive to lesion morphology, boundary sharpness, and surrounding tissue texture. Chest X-ray edits require the target sign to be radiographically meaningful without disrupting the global thoracic structure. Fundus edits must preserve vessel continuity, illumination, and local retinal texture while introducing the requested abnormality. These modality-dependent failure patterns justify storing verifier rationales at the trajectory level: the same rejection outcome can correspond to different visual or medical constraints, and therefore require different corrective prompts.
\end{document}